\documentclass[sigconf]{acmart}
\usepackage[super]{nth}
\usepackage{amsmath}
\usepackage{bm}
\usepackage{array}
\usepackage{makecell}
\usepackage{multirow}
\usepackage{tabularx}
\usepackage{graphicx}
\usepackage{caption}
\usepackage{subcaption}
\usepackage{url}
\usepackage{enumitem}
\usepackage{float}
\usepackage[flushleft]{threeparttable}

\AtBeginDocument{%
  \providecommand\BibTeX{{%
    \normalfont B\kern-0.5em{\scshape i\kern-0.25em b}\kern-0.8em\TeX}}}

\setcopyright{acmcopyright}
\copyrightyear{2018}
\acmYear{2018}
\acmDOI{10.1145/1122445.1122456}

\acmConference[Woodstock '18]{Woodstock '18: ACM Symposium on Neural
  Gaze Detection}{June 03--05, 2018}{Woodstock, NY}
\acmBooktitle{Woodstock '18: ACM Symposium on Neural Gaze Detection,
  June 03--05, 2018, Woodstock, NY}
\acmPrice{15.00}
\acmISBN{978-1-4503-XXXX-X/18/06}

\begin{document}
\setlength{\abovedisplayskip}{3.5pt plus 0pt}%
\setlength{\belowdisplayskip}{3.5pt plus 0pt}%
\title{Distance-wise Prototypical Graph Neural Network for Imbalanced Node Classification}


\author{Yu Wang}
\email{yu.wang.1@vanderbilt.edu}
\affiliation{%
  \institution{Vanderbilt University}
  \country{}
}

\author{Charu Aggarwal}
\email{charu@us.ibm.com}
\affiliation{%
  \institution{IBM T.J. Watson Research Center}
  \country{}
}

\author{Tyler Derr}
\email{tyler.derr@vanderbilt.edu}
\affiliation{%
  \institution{Vanderbilt University}
  \country{}
}



\begin{abstract}
Recent years have witnessed the significant success of applying graph neural networks (GNNs) in learning effective node representations for classification. However, current GNNs are mostly built under the balanced data-splitting, which is inconsistent with many real-world networks where the number of training nodes can be extremely imbalanced among the classes. Thus, directly utilizing current GNNs on imbalanced data would generate coarse representations of nodes in minority classes and ultimately compromise the classification performance. This therefore portends the importance of developing effective GNNs for handling imbalanced graph data. In this work,  we propose a novel Distance-wise Prototypical Graph Neural Network (DPGNN), which proposes a class prototype-driven training to balance the training loss between majority and minority classes and then leverages distance metric learning to differentiate the contributions of different dimensions of representations and fully encode the relative position of each node to each class prototype. Moreover, we design a new imbalanced label propagation mechanism to derive extra supervision from unlabeled nodes and employ self-supervised learning to smooth representations of adjacent nodes while separating inter-class prototypes. Comprehensive node classification experiments and parameter analysis on multiple networks are conducted and the proposed DPGNN almost always significantly outperforms all other baselines, which demonstrates its effectiveness in imbalanced node classification. The implementation of DPGNN is available at
\url{https://github.com/YuWVandy/DPGNN}.
\end{abstract}


\maketitle

\section{Introduction}\label{sec-introduction}
Graph neural networks (GNNs) have become one of the most promising paradigms in graph representation learning for node classification~\cite{GCN}, where a classifier is trained by labeled nodes and then used for categorizing labels of all remaining nodes. Although many GNN variants have been proposed to complete this task~\cite{hamilton2020graph, GAT, TDGNN}, 
prevailing prior works follow a (semi-)supervised setting where labeled nodes are assumed to be balanced among different classes~\cite{zhou2019meta}. However, this setting requires sufficient balanced labeled nodes for each class, which is over-idealized and inconsistent with reality.

In many real-world networks, the class distribution of labeled nodes is inherently skewed~\cite{graphproto, graphsmote, shi2020multi} where a large portion of classes (minority classes) only contain a limited number of labeled nodes (minority nodes) while few classes (majority classes) contain enough labeled nodes (majority nodes). Since most GNNs are designed without considering the potential of class imbalance, directly using them on imbalanced dataset would undermine the learned representations of minority nodes. The imbalanced node classification with GNNs naturally inherits existing challenges of deep learning in imbalanced classification: the inclination to learning towards majority classes~\cite{suervey} and the catastrophic forgetting of previous learned instances in minority classes~\cite{matching}. First, deep learning models improve representations for completing target tasks via backpropagation from the training loss. However, in the class imbalanced scenario, the main component of the training loss comes from majority classes and thus the gradient is dominated by majority classes such that the model is updated towards behaving significantly better on majority classes than minority ones. Second, deep learning models are notorious for demanding big 
data for updating parameters~\cite{matching}, which limits their ability to learn from minority classes that may only have a few training instances in the imbalanced setting. In addition, over-smoothing~\cite{Oversmoothing, Delioversmoothing} that occurs as a general issue in GNNs would become even worse in the imbalanced setting: representations of minority nodes would become similar to the majority ones and deviate from their own spectrum due to the imbalance bias introduced in the message passing.

Traditional methods for handling imbalance are either augmenting data via under(over)-sampling~\cite{chawla2004special}, or assigning weights to adjust the portion of training loss from different classes~\cite{thai2010cost}. Most of these methods are proposed for non-graph structured data. DR-GCN~\cite{shi2020multi} is the pioneer to explore node imbalance classification by adversarial training each class distribution and enforce the consistency between the labeled and unlabeled data distributions. However, the adversarial training may yield unrepresentative minority class distribution given few minority nodes~\cite{dong2019margingan}. Further, RECT~\cite{RECT} is proposed by leveraging topological regularization to derive extra supervision for minority nodes while it is designed for completely-imbalanced data and its performance on partially imbalanced data is unclear. GraphSMOTE~\cite{graphsmote} generalizes SMOTE~\cite{smote} to the graph domain by pre-training an edge generator and hence adding relational information for the new synthetic nodes from SMOTE. However, the computation  
of calculating the similarity between all pairs of nodes and pre-training the edge generator is extremely heavy. 

To tackle the aforementioned challenges of imbalanced node classification, we present Distance-wise Prototypical Graph Neural Network (DPGNN), which first applies class prototype-driven training to balance the training loss of different classes and then leverages distance metric learning to differentiate the contribution of each dimension of distance from each query node to all class prototypes. Ultimately the classification is performed by comparing the similarity of the learned distance metric representations of the nodes with the ones of class prototypes. Additionally, DPGNN introduces a novel imbalance label propagation scheme to augment training data and employs self-supervised learning to smooth representations of adjacent nodes while separating inter-class prototypes. The main contributions are summarized as follows:
\vspace{-1ex}
\begin{itemize}[leftmargin=0.5cm]
    \item We construct a balanced training scheme inspired by episodic training and further introduce distance metric learning to better capture node distances to class prototypes.

    \item We design a new imbalanced label propagation scheme to augment the training data and employ self-supervised learning to further improve the distance metric representations.
    
     \item We perform extensive imbalanced node classification experiments on real-world datasets across various levels of class imbalance with detailed parameter analysis to corroborate the effectiveness of our model. 
     
\end{itemize}

\vspace{-1.75ex}
\section{Related Work}\label{sec-relatedwork}
\subsection{Class Imbalance Problem}
Class imbalance exists in many real-world applications~\cite{he2013imbalanced}, where classes with more (or less) training instances are termed as majority (or minority) classes. The imbalance in the number of training instances among different classes significantly affects the performance of supervised learning and hence has become a classical research direction~\cite{suervey}. Generally, the approaches against this problem are summarized into three levels, i.e., data-level, algorithm-level and hybird~\cite{krawczyk2016learning}. Data-level methods aim to improve the data by balancing the training instances (e.g., up(down)-sampling~\cite{chawla2004special}), whereas algorithm-level methods~\cite{thai2010cost} attempt to improve the training process by modifying the learning algorithm such as re-weighting. In this work, we research on class imbalance in the graph domain. In the algorithm-level, our work balances the training loss by comparing labeled nodes with each class prototypes. In the data-level, we propose an imbalanced label propagation 
to obtain additional minority samples. Thus, 
our DPGNN is a hybrid approach.

\vspace{-0.5ex}
\subsection{Graph Neural Networks}
GNNs have achieved unprecedented success on graph-structured data due to the combination of feature propagation and prediction~\cite{rong2020deep}. However, most prior work on GNNs fails to consider the class imbalance problem, which unfortunately widely exists in real-world applications~\cite{graphsmote,RECT}. RECT~\cite{RECT} was developed merging a GNN and proximity-based embedding component for the completely-imbalanced setting (i.e., where some classes can even have no labeled nodes). DR-GCN~\cite{shi2020multi} explored node imbalance classification by adversarial training each class distribution and enforce the consistency between the labeled and unlabeled data distributions. Unfortunately, the adversarial trained generator and discriminator maybe overfitted to few minority nodes and hence loses generalizability~\cite{dong2019margingan}. More recently, GraphSMOTE~\cite{graphsmote} was designed using a GNN encoder to learn node embeddings and an extra edge generator to generate edges connecting synthetic minority nodes. However, the model is time-consuming and somewhat learned in an ad-hoc fashion. Motivated by this and to provide a grounded approach for imbalanced node classification, we design a novel GNN-based framework that utilizes class prototypes to balance the training loss and distance metric learning to fully encode the relative position of each node to each class prototype.

\vspace{-1ex}


\section{Problem Statement}\label{sec-probstatement}
We denote an attributed graph by $\mathcal{G} = (\mathcal{V}, \mathcal{E}, \mathbf{X})$ where $\mathcal{V} = \{v_1, ..., v_n\}$ is the set of $n$ nodes, $\mathcal{E}$ is the set of $m$ edges with $e_{ij}$ being the edge between nodes $v_i$ and $v_j$, and $\mathbf{X}=[\mathbf{x}_1^{\top}, ..., \mathbf{x}_n^{\top}] \in \mathbb{R}^{n\times d}$ is the node feature matrix with $\mathbf{x}_i$ indicating the features of node $v_i$. The network topology is described by its adjacency matrix $\mathbf{A}\in \{0, 1\}^{n\times n}$, where $\mathbf{A}_{ij} = 1$ denotes an edge between nodes $v_i$ and $v_j$, and $\mathbf{A}_{ij} = 0$ otherwise. The diagonal matrix of node degrees is denoted as $\mathbf{D}\in\mathbb{R}^{n\times n}$, where $\mathbf{D}_{ii} = \sum_{j}{\mathbf{A}_{ij}}$ calculates the degree of the node $v_i$. We let $\widetilde{\mathbf{A}} = \mathbf{A} + \mathbf{I}$ represent the adjacency matrix with added self-loops and similarly let $\widetilde{\mathbf{D}}=\mathbf{D} + \mathbf{I}$. Then the normalized adjacency matrix can be defined as $\widehat{\mathbf{A}} = \widetilde{\mathbf{D}}^{-0.5}\widetilde{\mathbf{A}}\widetilde{\mathbf{D}}^{-0.5}$. The neighborhood node set of the center node $v_i$ is given by $\mathcal{N}_{i} = \{v_j|e_{ij}\in\mathcal{E}\}$. Finally, $\mathcal{C} = \{1, 2, ..., C\}$ is the set of $C$ classes and $\mathbf{Y}\in\mathbb{R}^{n\times C}$ is the one-hot node label matrix. In Appendix~\ref{notation} we summarize main notations used throughout the paper and further explanations will be given in later sections. Now, given the previously defined notations, the imbalanced node classification can be mathematically formalized as follows: 

\textit{Given an attributed network $\mathcal{G} = (\mathcal{V}, \mathcal{E}, \mathbf{X})$ with labels for a subset of nodes $\mathcal{V}_l \subset	\mathcal{V}$ that are imbalanced among the $C$ 
classes, we aim to learn a node classifier $f: f(\mathcal{V}, \mathcal{E}, \mathbf{X}) \rightarrow \mathbf{Y}$ that can work well for both majority and minority classes.}

\begin{figure*}[ht!]
     \centering
     \includegraphics[width=.94\textwidth]{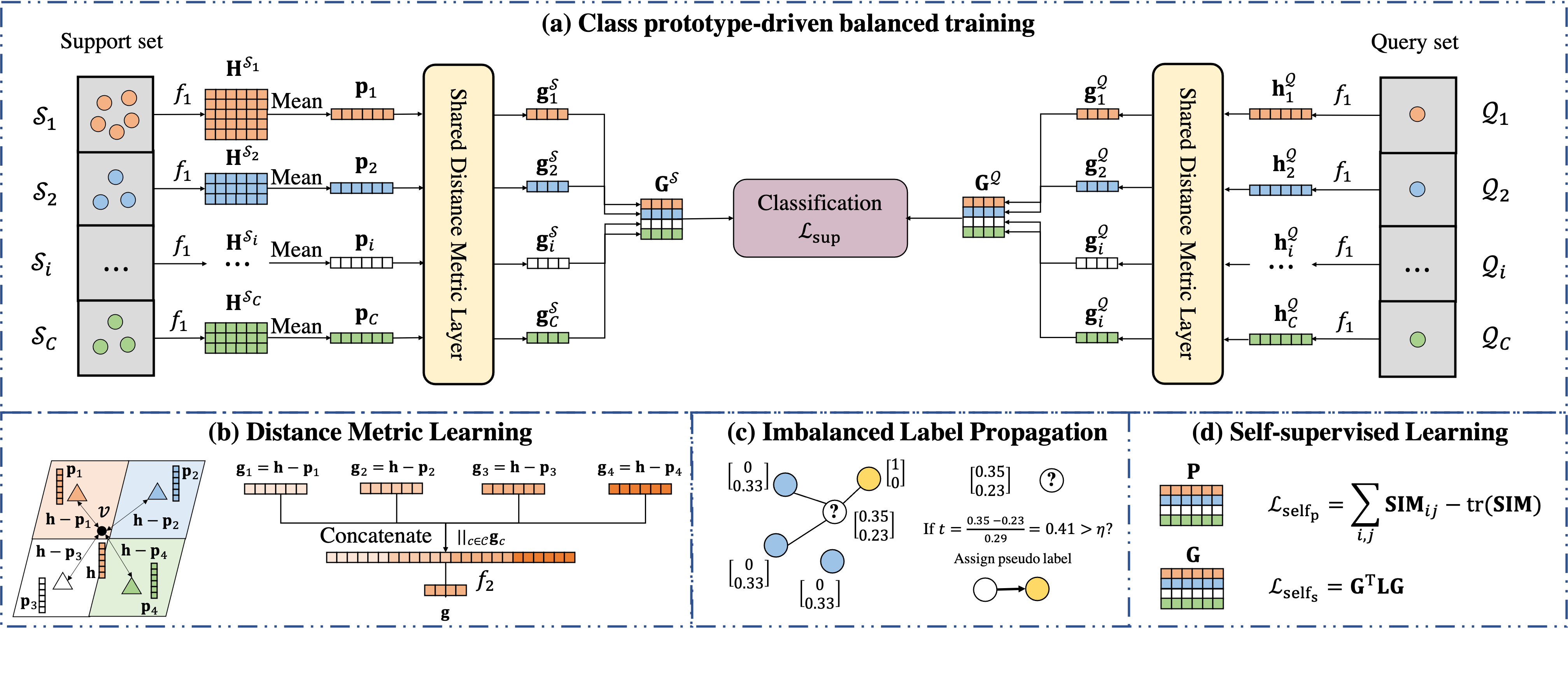}
     \vskip -5.5ex
     \caption{Overview of the Distance-wise Prototypical Graph Neural Network, with 
     four main components: (a) class prototype-driven balanced training, (b) distance metric learning, (c) imbalanced label propagation, and (d) self-supervised learning.}
     \label{fig-framework}
     \vspace{-2ex}
\end{figure*}

\vspace{-0.75ex}
\section{The proposed framework}\label{sec-framework}
In this section, we present our proposed Distance-wise Prototypical Graph Neural Network (DPGNN) that is able to solve the challenges imposed on deep graph learning when given imbalanced training data. The framework is shown in Figure~\ref{fig-framework}, including four main components: (a) class prototype-driven balanced training, (b) distance metric learning, (c) imbalanced label propagation, and (d) self-supervised learning. Specifically, we borrow the idea episodic training from prototypical networks~\cite{proto} to first balance the training loss across different classes via prototype-driven balanced training (Figure~\ref{fig-framework}(a)). However, unlike traditional prototypical networks where labels of query nodes are assigned based on their nearest class prototypes, we further employ distance metric learning to construct another distance metric space that can differentiate different distance dimensions and fully characterize the position of each query node relative to all class prototypes (Figure~\ref{fig-framework}(b)). In order to fully incorporate the graph topology information in alleviating the imbalance problem, we also design a novel imbalanced label propagation scheme and two self-supervised components to aid in learning high-quality distance metric representations (Figure~\ref{fig-framework}(c) and ~\ref{fig-framework}(d), respectively).  Next we describe each 
component in detail.

\subsection{Class Prototype-driven Balanced Training}\label{sec-episodictraining}
In order to balance the training loss from majority and minority classes, we leverage the idea of episodic training by sampling support and query sets, and then calculating representations of prototypes for each class.  In order to avoid using the original sparse and high-dimensional node features and to allow learned complex feature transformation, we first apply a GNN-based encoder $f_1: \mathbb{R}^{n\times d}\times \mathbb{R}^{n\times n} \rightarrow \mathbb{R}^{n\times d'}$ to obtain $d'$-dimensional node representations $\mathbf{H} \in \mathbb{R}^{n\times d'}$. Most GNN-based encoders $f_1$ can be decomposed into two components: neighborhood propagation and feature transformation, which can be generally formalized as:
\begin{equation}\label{eq-propagation}
    \mathbf{h}_i^l = \text{TRAN}^{l}(\text{AGGR}^{l}(\mathbf{h}_{i}^{l - 1}, \text{PROP}^{l}(\{\mathbf{h}_j^{l - 1}|v_j\in\mathcal{N}_{i}\}))),
\end{equation}
where in each layer $l$, the neighborhood representations $\{\mathbf{h}^{l - 1}_{j}|v_j\in\mathcal{N}_{i}\}$ are first propagated via $\text{PROP}^{l}$ to node $v_i$ and aggregated with its own representation $\mathbf{h}_i^{l - 1}$ by $\text{AGGR}^{l}$, then the combined representation is further transformed by $\text{TRAN}^{l}$ to output the representation $\mathbf{h}_i^{l}$ of node $v_i$ after layer $l$ of the GNN. 

With the learned node embeddings $\mathbf{H}$ coming from the GNN-based encoder's last layer, we aim to compute representations of class prototypes $\mathbf{P}\in\mathbb{R}^{C\times d'}$. During each training epoch, the model is fed with an episode sampled from the labeled training nodes $\mathcal{V}_l$, which is further divided into support sets $\mathcal{S} = \{\mathcal{S}_c|c\in\mathcal{C}\}$ and query sets $\mathcal{Q} = \{\mathcal{Q}_c|c\in\mathcal{C}\}$ of each class. The query sets only contain one training sample from each class and provide a balanced way to train the encoder. The support sets are formed by grouping the remaining training samples from each class except the query samples from the query sets, which serves as anchors to characterize the class prototypes of each class. Following the idea of Prototypical Networks~\cite{proto}, we define class prototypes to be closely surrounded by nodes of the same class, such that they can precisely represent their class. More specifically, the prototype $\mathbf{p}_c$ of class $c$ is computed by:
\begin{equation}\label{eq-proto}
    \mathbf{p}_c = \text{PROTO}(\mathbf{H}^{\mathcal{S}_c}) = \text{PROTO}(\{\mathbf{h}_i| v_i\in\mathcal{S}_c\}), c\in\mathcal{C},
\end{equation}
where $\text{PROTO}$ is the prototype computation which calculates the representation of a class prototype $\mathbf{p}_c$ based on node representations $\mathbf{H}^{\mathcal{S}_c}$ that come from the support set $\mathcal{S}_c$. For instance, in the vanilla Prototypical Network~\cite{proto}, the mean-pooling is employed here:
\begin{equation}\label{eq-meanpooling}
    \mathbf{p}_c = \frac{1}{|\mathcal{S}_c|}\sum_{v_i\in\mathcal{S}_c}{\mathbf{h}_i}, c\in\mathcal{C}.
\end{equation}
Applying mean-pooling to calculate class prototypes assumes that each class can be represented using only one class prototype, which might be underrepresentative when a unimodal distribution assumption is violated. Thus, multi-prototypes to represent each class could be used, such as replacing mean-pooling with another permutation-invariant function 
(e.g., K-means clustering~\cite{hartigan1979algorithm}), or even iteratively learn multi-prototypes 
according to the complexity of the class distribution~\cite{allen2019infinite}. We leave this as one future direction.

Prototypes serve as representatives of each class and can be used to classify query nodes by selecting their nearest prototypes. However, directly employing a softmax over distances to prototypes in the embedding space to obtain the class probability distribution~\cite{matching, proto, graphproto} will make different dimensions of distance contribute equally to the classification
and in high dimensional data pairwise Euclidean distances tend to converge~\cite{aggarwal2001surprising}. Furthermore, using only 
embedding distance of a node to the nearest prototype omit its embedding distance to all other non-nearest prototypes that may encode extra information about the node's class. Thus, instead of classifying query nodes directly based on their nearest prototype~\cite{graphproto, proto}, we devise a distance metric layer to project nodes from the original embedding space to another distance metric space, where query nodes are classified by comparing their learned distance metric representations with the ones of class prototypes.

\vspace{-0.25ex}
\subsection{Distance Metric Learning}
In order to project each node from the original embedding space to the distance metric space, we first concatenate the difference of its embedding from each class prototype. Next we apply a linear transformation on top of that, to pay different attention on each original distance dimension and adaptively extract useful distance information. Figure~\ref{fig-framework}(b) demonstrates the detailed procedure of computing distance metric representation $\mathbf{g}$ of node $v$.

Given a node $v$ with embedding $\mathbf{h}\in\mathbb{R}^{d'}$, we calculate its distance metric representation $\mathbf{g}_c\in\mathbb{R}^{d'}$ to each class prototype $c$ as:
\begin{equation}\label{eq-diff}
    \mathbf{g}_c = \mathbf{h} - \mathbf{p}_c, c\in\mathcal{C},
\end{equation}
where $\mathbf{h} - \mathbf{p}_c$ calculates the difference of the embedding between each node and each class prototype. For each node, only considering its embedding difference to one class prototype cannot fully locate its position. Therefore, we concatenate the difference of the node's embedding to all class prototypes $||_{c\in\{1, ..., C\}}\mathbf{g}_c$ and further apply a linear transformation $f_\text{2}: \mathbb{R}^{d'C} \rightarrow \mathbb{R}^{d''}$ 
to pay different levels of attention to different dimensions of the embedding difference and adaptively extract useful embedding difference information.
\begin{equation}\label{eq-concat}
    \mathbf{g} = f_2(||_{c\in\mathcal{C}}\mathbf{g}_c).
\end{equation}
The distance metric representation $\mathbf{g}$ encodes the distance information of the node $v$ to all class prototypes, which as a result precisely captures its relative position to all class prototypes. In order to use these distance metric representations to prototypes for reference to classify query nodes, we feed the representations of prototypes and query nodes $\mathbf{p}_c, \mathbf{h}_c^{\mathcal{Q}},$ for $c\in \mathcal{C}$ into the shared distance metric layer to learn their distance metric representations $\mathbf{g}^{\mathcal{S}}_c, \mathbf{g}^{\mathcal{Q}}_c,$  for $c\in\mathcal{C}$. Then we stack them across all classes to compute the distance metric representations of prototypes and query nodes:
\begin{equation}\label{eq-stackS}
    \mathbf{G}^{\mathcal{M}} = \text{stack}(\mathbf{g}^{\mathcal{M}}_c | c\in\mathcal{C}), \mathcal{M} \in \{\mathcal{Q}, \mathcal{S}\}
\end{equation}
Next, the predicted class distribution for each query node is:
\begin{equation}
    \mathbf{F} = \text{softmax}(\mathbf{G}^{\mathcal{Q}}(\mathbf{G}^{\mathcal{S}})^{\top}),
\end{equation}
where $\mathbf{F}_c$ gives the predicted probability distribution of class $c$'s query node over all classes, which is then used to calculate the supervised classification loss:
\vspace{-1.5ex}
\begin{equation}
    \mathcal{L}_\text{class} = \frac{1}{C}{\sum_{c = 1}^{C}\ell(\mathbf{F}_c, c)},
\end{equation}
where $\ell(\cdot, \cdot)$ is a loss function to measure the difference between predictions and ground-truth labels, such as cross entropy.


\vspace{-1ex}
\subsection{Imbalanced Label Propagation}
Network homophily~\cite{mcpherson2001birds} assumes that connected nodes tend to share similar attributes or belong to the same class, which is commonly the case in various real-world networks~\cite{coin, zhu2020beyond}. Thus, it be can naturally harnessed to augment the training data especially for increasing the minority training nodes. Motivated by this, we perform imbalanced label propagation to annotate node labels based on their neighboring nodes' labels and filter out unconfident ones by computing their topology information gain~\cite{chen2020distance}. 

Since we only have access to the class information of labeled nodes $\mathcal{V}_l$ and to balance the voting effect between the majority and the minority classes, for unlabeled nodes we mask their one-hot labels by filling zero vector $\mathbf{0} \in \mathbb{R}^{C}$, and for labeled nodes we multiply their one-hot labels by the weighting factor $\gamma_i$:
\begin{equation}
    \widetilde{\mathbf{Y}}_i = \begin{cases}
    \gamma_i\mathbf{Y}_i, &v_i\in\mathcal{V}_l\\
    \mathbf{0}, &v_i\in\mathcal{V}\backslash \mathcal{V}_l,
    \end{cases}
    \text{~where~} \gamma_i = \frac{|\mathcal{V}_l|}{\sum_{v_j\in\mathcal{V}_l}\mathbf{Y}_{j\phi(v_i)}}
\end{equation}
computes the inverse ratio of the number of labeled nodes in the node $v_i$'s class $\phi(v_i), \sum_{v_j\in\mathcal{V}_l}\mathbf{Y}_{j\phi(v_i)}$, to the total number of labeled nodes $|\mathcal{V}_l|$. Then we propagate this unbiased label distribution $\widetilde{\mathbf{Y}}$ to their neighbors that are at most 
$k$-hops away by considering $k^{\text{th}}$-order adjacency matrix as follows: 
\begin{equation}\label{eq-kprop}
    \widehat{\mathbf{Y}} = \widehat{\mathbf{A}}^{k}\widetilde{\mathbf{Y}}.
\end{equation}
Note that $\widehat{\mathbf{A}}^{k}\widetilde{\mathbf{Y}}$ can be computed efficiently by applying power iteration sequentially from right to left~\cite{APPNP}. 

However, this leads to another issue that 
$\widehat{\mathbf{Y}}$ cannot be used directly in 
class prototype-driven balance training to sample support and query sets since $\widehat{\mathbf{Y}}$ is a soft-label (instead of having hard label assignments). Moreover, nodes that fall in the topological boundary between different classes may possess noisy label distribution and mislead the training process. Therefore, we utilize the idea of topological information gain (TIG)~\cite{chen2020distance} to filter out nodes with weak and obscure label distributions. TIG describes the task information effectiveness that the node obtains from labeled source along the network topology~\cite{chen2020distance}. We regard the maximum entry of the soft-label $\widehat{\mathbf{Y}}_i$ to be the most possible class type that the node $v_i$ can be and the other entries as confusing information. Then the topological information gain $\mathbf{t}_i$ for node $v_i$ is calculated as:
\begin{equation}
    \mathbf{t}_i = \frac{\max(\widehat{\mathbf{Y}}_i) - \frac{(\sum_{c\in \mathcal{C}}{\widehat{\mathbf{Y}}_{ic} - \max(\widehat{\mathbf{Y}}_i)})}{C - 1}}{\frac{1}{C}\sum_{c\in\mathcal{C}}{\widehat{\mathbf{Y}}_{ic}}}.
\end{equation}
A high $\mathbf{t}_i$ means the label distribution of node $v_i$ is sharp/strong and thus it lies in the cluster of nodes in class $\text{argmax}(\widehat{\mathbf{Y}}_i)$. By network homophily, its label is also conjectured to be $\text{argmax}(\widehat{\mathbf{Y}}_i)$. As such, we generate the hard pseudo label of non-training node $v_i$ by binary thresholding its topological information gain $\mathbf{t}_i$ as:
\begin{equation}\label{eq-threshold}
    \check{\mathbf{Y}}_{ic} = \begin{cases}
    1&, \text{if $\mathbf{t}_i > \eta$ and $c=\text{argmax}(\widehat{\mathbf{Y}}_i)$}\\
    0&, \text{if $\mathbf{t}_i\le\eta$ or $c\ne \text{argmax}(\widehat{\mathbf{Y}}_i)$}\\
    \end{cases}, v_i\in\mathcal{V}\backslash \mathcal{V}_l,
\end{equation}
where $\eta$ is a hyperparameter that controls the trade-off between the quality and the number of the augmented labels. Higher $\eta$ leads to precise labels with sharp distribution and thus guarantee the labeling quality but less nodes could be augmented. Lower $\eta$ leads to more imprecise and noisy labels with even distribution. Note that for the initial labeled nodes $v_i\in\mathcal{V}_l$, we still utilize their original labels and thus the final labels we use for sampling support and query sets in class prototype-driven balanced training:
\begin{equation}
    \bar{\mathbf{Y}}_i = \begin{cases}
    \mathbf{Y}_i & v_i \in \mathcal{V}_l\\
    \check{\mathbf{Y}}_i & v_i \in \mathcal{V}\backslash \mathcal{V}_l\\
    \end{cases}
\end{equation}
Details of hyperparameter tuning $\eta$ are provided in Section~\ref{sec-etatune}.

\vspace{-2ex}
\subsection{Self-Supervised Learning (SSL)}
Although node embeddings computed from the GNN-based encoder $f_1$ embed network topology information, the prototype computation $\text{PROTO}$ in Eq.~\eqref{eq-proto} and the distance metric learning in Eq.~\eqref{eq-diff} do not consider network topology. Inspired by the intuition that different class prototypes should have different representations and adjacent nodes should have similar distance metric representations, we design two 
GNN self-supervised learning (SSL)~\cite{wangself} 
pretext tasks to emphasize the topological information learned by DPGNN.

To ensure that different prototypes have different representations, we minimize the representation similarity of prototypes from different classes as follows:
\begin{equation}\label{eq-lossselfp}
\small
    \mathcal{L}_{\text{ssl}_\text{p}} = \sum_{i=1}^{C}{\sum_{j = 1}^{C} \big( \mathbf{SIM}_{ij}} \big) - \text{tr}(\mathbf{SIM}),
    \text{~where~}
    \mathbf{SIM}_{ij} = \frac{{\mathbf{p}_i}^{\top}\mathbf{p}_j}{||\mathbf{p}_i|| \hspace{0.25ex} ||\mathbf{p}_j||}
\end{equation}

To smooth the distance metric representations between adjacent nodes and reinforce the graph structure in the learned representations,
we adopt the following objective function:
\begin{equation}\label{eq-lossselfs}
    \mathcal{L}_{\text{ssl}_\text{s}} = \mathbf{G}^{\top}\mathbf{L}\mathbf{G},
\end{equation}
where $\mathbf{G}$ is the distance metric representations by transforming the concatenated embedding distance between each node to each class prototype (i.e., Eq.~\eqref{eq-diff} and Eq.~\eqref{eq-concat}), and $\mathbf{L}$ is the Laplacian matrix of the underlying network. If we adopt the normalized Laplacian matrix, i.e., $\mathbf{L} = \mathbf{I} - \widehat{\mathbf{A}}$, Eq.~\eqref{eq-lossselfs} can be rewritten as:
\begin{equation}\label{eq-smooth}
    \mathcal{L}_{\text{ssl}_\text{s}} = \sum_{v_i\in\mathcal{V}}{\sum_{v_j\in\mathcal{N}_{i}}{\Big(\frac{\mathbf{g}_i}{\sqrt{d_i}} - \frac{\mathbf{g}_j}{\sqrt{d_j}}\Big)^2}}.
\end{equation}
It is clear that $\mathcal{L}_{\text{ssl}_\text{s}}$ is small when adjacent nodes share similar distance metric representations to class prototypes.

Collecting one supervised classification loss and two SSL loss, the overall objective function is stated as:
\begin{align}
    \min{\mathcal{L}} &= \mathcal{L}_{\text{class}} + \lambda_{1}\mathcal{L}_{{\text{ssl}_\text{p}}} + \lambda_{2}\mathcal{L}_{{\text{ssl}_\text{s}}},
\end{align}
where $\lambda_{1}$ and $\lambda_{2}$ are two hyperparameters that control the contribution of the two SSL losses (i.e., $\mathcal{L}_{\text{ssl}_\text{p}}$ and  $\mathcal{L}_{\text{ssl}_\text{s}}$) 
in addition to the supervised classification loss 
$\mathcal{L}_{\text{class}}$.

\vspace{-0.5ex}
\subsection{Complexity Analysis}
Having introduced all components of DPGNN, next we compare DPGNN with vanilla GNN-based encoders by analyzing the additional complexity in terms of time and model parameters.

In comparison to vanilla GNN-based encoders, additional computational requirements come from three components: distance metric learning, imbalanced label propagation, and self-supervised learning. For distance metric learning, the calculation of the pairwise embedding difference between each node and each class prototype requires $O(|\mathcal{V}|C)$ time complexity if implemented na\"ively. Typically the class number $C$ is multiple orders of magnitude less than the node number $|\mathcal{V}|$ in a network and therefore can be treated as a constant, which leads to linear time complexity $O(|\mathcal{V}|)$. For uncommon networks that have an extremely large number of classes with the same magnitude as the size of the network, we could select sub-classes $\mathcal{C}_{\text{sub}}$ as anchors and approximate the distance metric representations by considering distance to these anchors rather than all classes~\cite{liu2010large, wu2019scalable, chen2020iterative}. In imbalanced label propagation, the most computational part comes from propagating labels in Eq.~\eqref{eq-kprop}, which can be completed efficiently by applying power iteration from the edge view in $O(k|\mathcal{E}|)$ compared to $O(|\mathcal{V}|^3)$ for matrix calculation of $\widehat{\mathbf{A}}^k$. Among the two SSL components, the heaviest computation comes from smoothing distance metric embeddings of adjacent nodes in Eq.~\eqref{eq-lossselfs}, 
which can 
be calculated from the edge view as Eq.~\eqref{eq-smooth}; thus, 
is linear with 
the number of edges $O(|\mathcal{E}|)$.

For the model complexity, apart from the parameters of GNN-based encoder, additional parameters of DPGNN come from the linear transformation $f_2$, which are $\mathbf{W}^{f_2}$ and $\mathbf{b}^{f_2}$. Hence, compared with vanilla GNN-based encoders, the overall additional parameters 
are $O(d'C\times d'')$ where $d'C$ is the dimension of concatenated distance metric representations to all class prototypes $||_{c\in\{1, ..., C\}}\mathbf{g}_c$ and $d''$ is the dimension of distance metric representation $\mathbf{g}$ (as in Eq.~\eqref{eq-concat}). Typically, if the number of classes $C$ is  
small or if we only select some sub-classes as anchors, $d'C$ will be far less than the original high-dimensional node attributes $d$. Therefore, the extra model complexity $O(d'C\times d'')$ could practically be ignored compared to the complexity of the GNN-based encoder, which has $O(d\times d')$.

\section{Experimental Results}\label{sec-experiment}
In this section, we conduct extensive experiments on imbalanced node classification to evaluate the effectiveness of DPGNN. In particular, we target to answer the following three questions:
\begin{itemize}[leftmargin=0.5cm]
    \item \textbf{Q1:} How effective is DPGNN compared to other baselines on 
    imbalanced node classification under different imbalance ratio?
    
    \item \textbf{Q2:} How do different components of DPGNN contributes to performance improvement in imbalanced node classification?
    
    \item \textbf{Q3:} How does the threshold $\eta$ in imbalanced label propagation affect the performance of DPGNN? 
\end{itemize}
\vspace{-2ex}

\subsection{Experiment Settings}
\subsubsection{Datasets}
We experiment on widely-adopted citation networks~\cite{yang2016revisiting}, Amazon product networks~\cite{shchur2018pitfalls}, and an online social network~\cite{rozemberczki2021multi}. Table~\ref{tab-dataset} presents the basic network statistics for these datasets. In the five citation networks and the online social network, class distributions are relatively balanced, so we use an imitative imbalanced setting: we choose the first 5, 4, 2, 5, 3, 1 class(es) as minority and down-sample their training nodes to 2 compared to 20 for other majority class(es), which creates an imbalanced class distribution with imbalance ratio 10. For the Amazon product networks 
whose class distributions are genuinely imbalanced, we use their original class ratios and set the total training nodes as 50 and 30, respectively. For validation and testing sets, 500 and 1000 nodes are selected respectively for all 8 datasets, which is commonly employed in the literature~\cite{GCN, GAT}. This setting is used throughout the paper unless otherwise stated.

\begin{table}[t]
\footnotesize
\setlength{\extrarowheight}{.095pt}
\setlength\tabcolsep{3pt}
\caption{Basic dataset statistics.}
\vspace{-4ex}
\centering
\begin{tabular}{lcccccc}
 \Xhline{2\arrayrulewidth}
\textbf{Networks} & \textbf{Nodes} & \textbf{Edges} & \textbf{Features} & \textbf{Classes} & \textbf{Homophily}$^*$ & \textbf{Type} \\
 \Xhline{1.5\arrayrulewidth}
Cora & 2,708 & 5,429 & 1,433 & 7 & 0.81 & Citation\\
Citeseer & 3,327 & 4,732 & 3,703 & 6 & 0.74 & Citation\\
Pubmed & 19,717 & 44,338 & 500 & 3  & 0.80 & Citation\\
Cora-ML & 2,995 & 8,416 & 2,879 & 7  & 0.79 & Citation\\
DBLP & 17,716 & 105,734 & 1,639 & 4  & 0.83 & Citation\\
\makecell[l]{Amazon\\~Computers} &  13,381 & 245,778 & 767 & 10 & 0.78 & \makecell[c]{Online\\Product}\\
\makecell[l]{Amazon\\~Photo}  &  7,487 & 119,043 & 745 & 8  & 0.83 & \makecell[c]{Online\\Product}\\
Twitch PT & 1,912 & 64,510 & 128 & 2  & 0.58 & Social\\
 \Xhline{2\arrayrulewidth}
\end{tabular}
\label{tab-dataset}
\begin{tablenotes}
      \footnotesize
      \centering
      \item \textbf{*} Definition of homophily is attached in Appendix~\ref{sec-homophily}.
\end{tablenotes}
\vskip -3ex
\end{table}

\begin{table*}[htbp!]
\footnotesize
\setlength{\extrarowheight}{.095pt}
\setlength\tabcolsep{3pt}
\caption{Node classification performance on eight datasets with the best performance emboldened and second underlined.}
\vspace{-2ex}
\begin{tabular}{|l|ccc|ccc|ccc|ccc|}
\hline
 & \multicolumn{12}{c|}{\textbf{Dataset} (Homophily Value) } \\
\cline{2-13}
\multirow{2}{*}{\textbf{Model}} & \multicolumn{3}{c|}{\textbf{Cora} (0.81)} & \multicolumn{3}{c|}{\textbf{Citeseer} (0.74)} & \multicolumn{3}{c|}{\textbf{Pubmed} (0.80)} & \multicolumn{3}{c|}{\textbf{Cora-ML} (0.79)} \\
\cline{2-13}
 & F1-macro & F1-weight & F1-micro & F1-macro & F1-weight & F1-micro & F1-macro & F1-weight & F1-micro & F1-macro & F1-weight & F1-micro \\
 \hline
GCN & 0.5205 & 0.5195 & 0.5212 & 0.3870 & 0.4169 & 0.4692 & 0.5501 & 0.5569 & 0.5928 & 0.5205 & 0.5195 & 0.5212 \\
GCN$_{\text{us}}$ & 0.5631 & 0.5659 & 0.5727 & \underline{0.4503} & \underline{0.4822} & \underline{0.5220} & \underline{0.6272} & \underline{0.6323} & \underline{0.6451} & 0.5656 & 0.5516 & 0.5611\\
GCN$_{\text{rw}}$ & 0.5609 & 0.5660 & 0.5724 & 0.4457 & 0.4800 & 0.5156 & 0.6169 & 0.6178 & 0.6327 & 0.5609 & 0.5660 & 0.5724 \\
GCN$_{\text{st}}$ & 0.5488 & 0.5398 & 0.5519 & 0.4462 & 0.4794 & 0.5127 & 0.5861 & 0.5964 & 0.6186 & 0.5488 & 0.5398 & 0.5519\\
GraphSMOTE & \underline{0.5845} & \underline{0.6026} & \underline{0.5820} & 0.4236 & 0.4774 & 0.5020 & 0.6122 & 0.5998 & 0.6110 & \underline{0.6233} & \underline{0.6450} & \underline{0.6130}\\
RECT & 0.5234 & 0.5025 & 0.5448 & 0.4002 & 0.4243 & 0.4549 & 0.5713 & 0.5597 & 0.6002 & 0.5530 & 0.5560 & 0.6026\\
DR-GCN &0.5513 & 0.5362 & 0.5520 & 0.3924 & 0.4414 & 0.4880 &0.5628 & 0.5730 & 0.5559 & 0.5412 & 0.4716 & 0.5060\\
DPGCN & \textbf{0.7115} & \textbf{0.7029} & \textbf{0.7111} & \textbf{0.4838} & \textbf{0.5180} & \textbf{0.5397} & \textbf{0.7018} & \textbf{0.7176} & \textbf{0.7189} & \textbf{0.7273} & \textbf{0.7278} & \textbf{0.7305}\\

\hline
\end{tabular}

\hspace{0ex}
\vspace{0.5ex}
\begin{tabular}{|l|ccc|ccc|ccc|ccc|}
\hline
\multirow{2}{*}{\textbf{Model}} & \multicolumn{3}{c|}{\textbf{DBLP} (0.83)} & \multicolumn{3}{c|}{\textbf{Amazon Computers} (0.78)} & \multicolumn{3}{c|}{\textbf{Amazon Photo} (0.83)} & \multicolumn{3}{c|}{\textbf{Twitch PT} (0.58)} \\
\cline{2-13}
 & F1-macro & F1-weight & F1-micro & F1-macro & F1-weight & F1-micro & F1-macro & F1-weight & F1-micro & F1-macro & F1-weight & F1-micro \\
 \hline
GCN & 0.3482 &  0.3829 & 0.3876 & 0.5343 & 0.6808 & 0.6975 & 0.6999 & 0.7617 & 0.7666 & 0.4557 & 0.4510 & 0.4656\\

GCN$_{\text{us}}$ & 0.4214 & 0.4599 & 0.4795 & 0.5757 & 0.6876 & 0.6883 & 0.7135 & 0.7645 & 0.7632 & 0.4917 & 0.5088 & 0.5131\\
GCN$_{\text{rw}}$ & 0.4379 & 0.4744 & \underline{0.4892} & 0.5732 & 0.6845 & 0.6841 & 0.7204 & 0.7683 & 0.7670 & 0.4963 & 0.5168 & 0.5193\\
GCN$_{\text{st}}$ & 0.3757 & 0.4302 & 0.4522 & \underline{0.5863} & 0.6999 & 0.7107 & 0.7302 & 0.7782 & 0.7800 & \underline{0.5002} & 0.5267 & 0.5301\\
GraphSMOTE & \underline{0.4844} & \underline{0.4938} & 0.4530 & 0.5509 & 0.6213 & 0.6370 & 0.7227 & 0.7716 & 0.7750 & 0.3922 & 0.3558 & 0.4130\\
RECT & 0.3438 & 0.3602 & 0.3810 & 0.5222 & \underline{0.7002} & \textbf{0.7351} & 0.6858 & 0.7763 & \underline{0.8007} & 0.4898 & 0.4843 & 0.4904 \\
DR-GCN & 0.3797 & 0.4190 & 0.4510 & 0.5357 & 0.6745 & 0.7100 & \underline{0.7434} & \textbf{0.7979} & \textbf{0.8130} & 0.4791 & \underline{0.5524} & \underline{0.5750}\\
DPGCN & \textbf{0.6167} & \textbf{0.6665} & \textbf{0.6597} & \textbf{0.6702} & \textbf{0.7280} & \underline{0.7310} & \textbf{0.7600} & \underline{0.7943} & 0.7917 & \textbf{0.5600} & \textbf{0.5944} & \textbf{0.5915}\\

\hline
\end{tabular}
\label{table-F1macro}
\vskip -2.5ex
\end{table*}

\vspace{-1ex}
\subsubsection{Baselines}
To evaluate the effectiveness of the proposed DPGNN framework, we select six representative approaches for handling imbalance classification including the current state-of-the-art methods, where the first three target at point-based imbalance classification while the last three are designed specifically for imbalance node classification in networks.

\begin{itemize}[leftmargin=0.5cm]
    \item \textbf{Up-sampling:} 
    A classical approach that repeats samples from minority classes. Following~\cite{graphsmote}, we implement this in the embedding space by duplicating minority nodes' representations.
    
    \item \textbf{Re-weight~\cite{yuan2012sampling+}:} 
    A cost-sensitive approach that assigns class-specific loss weights; we set the weights of each class as inverse ratio of the total training nodes to the number in that class.
    
    \item \textbf{SMOTE~\cite{smote}:} Synthetic minority samples are created by interpolating minority samples with their nearest neighbors within the same class based on the output of last GNN layer.
    
    \item \textbf{GraphSMOTE~\cite{graphsmote}:} 
    Advancing SMOTE~\cite{smote}, GraphSMOTE has two types of edge generators that can be pre-trained to connect synthetic nodes to the original graph. We report the result of the best generator variant for each dataset.
    
    \item \textbf{RECT~\cite{RECT}:} 
    A supervised model leveraging both a GNN and an unsupervised node proximity-based embedding, which is designed for completely-imbalanced label setting.
    
    \item \textbf{DR-GCN~\cite{shi2020multi}:} Two types of regularization to tackle class imbalance are proposed: class-conditional adversarial training to separate labeled nodes and unlabeled nodes latent distribution constraint to maintain training equilibrium.
    
\end{itemize}

Equipping the first three baselines with a 2-layer GCN~\cite{GCN} encoder, we then have 7 baselines: $\text{GCN}$, $\text{GCN}_{\text{us}}$, $\text{GCN}_{\text{rw}}$, $\text{GCN}_{\text{st}}$, $\text{GraphSMOTE}$, RECT, and DR-GCN that are used for a comprehensive empirical analysis on imbalanced node classification.

\vspace{-1ex}
\subsubsection{Evaluation Metrics}
Following existing work in 
imbalanced classification~\cite{gao2021setconv}, 
we 
measure performance with: F1-macro, F1-micro, 
and F1-weighted scores. 
F1-macro and F1-weighted scores evaluate model performance across different classes, 
with the former taking 
unweighted mean over the accuracy of each class and the latter 
weighted mean to account for label imbalance.
F1-micro is taken 
over all testing examples, which gives 
an overall evaluation of the 
performance while 
undervalues nodes in minority classes.

\vspace{-1ex}
\subsubsection{Parameter Settings}
We implement our proposed DPGNN and some necessary baselines Pytorch Geometric~\cite{paszke2019pytorch}. For GraphSmote\footnote{GraphSMOTE Code: \url{https://github.com/TianxiangZhao/GraphSmote}}, RECT\footnote{RECT Code: \url{https://github.com/zhengwang100/RECT}}, and DR-GCN\footnote{DR-GCN Code: \url{https://github.com/codeshareabc/DRGCN}}, we use the authors' original code with any needed modifications from their 
GitHub repositories. Aiming to provide a rigorous/fair comparison, 
we tune hyperparameters for all models individually on each dataset around the default/best settings reported in their paper.
For DPGCN, we tune the following hyperparameters: dropout rate $\in\{0, 0.5\}$, the coefficient balancing the loss contribution $\lambda_1, \lambda_2 \in \{1, 10\}$, the threshold $\eta \in [0, 6]$, and the training epoch $\in\{1000, 3000, 6000\}$. We select the 2-layer GCN with 256 hidden units as the encoder $f_1$ due to its simplicity and efficiency, and termed our framework as DPGCN in the following. Note that any other GNN-based encoder can be used instead. For reproducibility, the code of our model is publicly available\footnote{DPGNN Code: \url{https://github.com/YuWVandy/DPGNN}} and Appendix~\ref{app-hyper} has further hyperparameter configuration details. 

\begin{figure*}[htbp!]
     \centering
     \includegraphics[width=.92\textwidth]{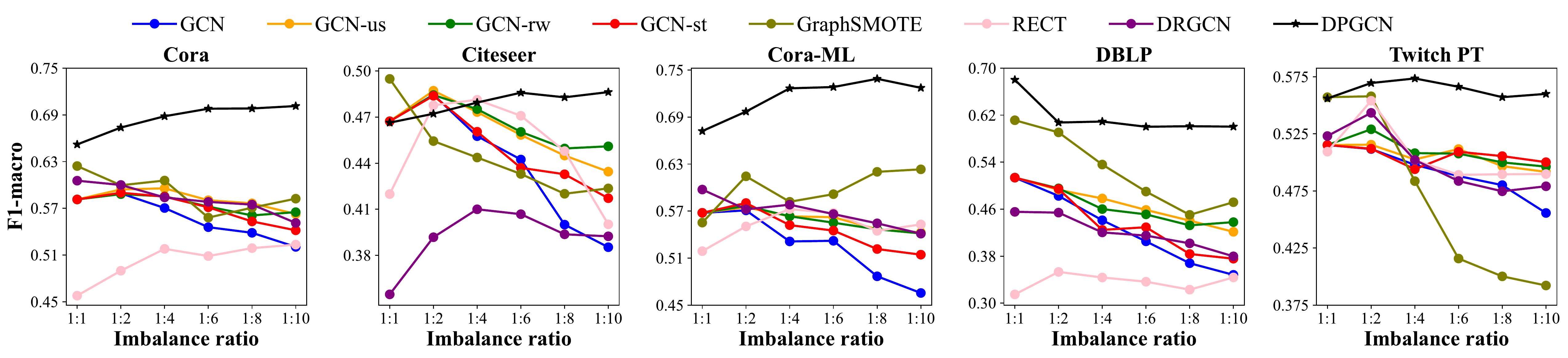}
     \vspace{-3ex}
     \caption{Node classification results under different class imbalance ratios.}
     \label{fig-imbratio}
     \vspace{-1.5ex}
\end{figure*}

\subsection{Performance Comparison}
In this subsection, we answer the first question by comparing the performance of DPGCN with other baselines and report the average perforance per metric across 20 different data splits along with the edge homophily~\cite{zhu2020beyond} for each dataset. We observe that DPGCN performs the best in all 8 datasets by F1-macro with a large margin, and best on 7(6) of the 8 datasets according to F1-weight (F1-micro) over other baselines. The F1-micro of GCN is higher than F1-macro on all datasets except Cora, which indicates accuracy is vastly different across different classes and therefore signifies the detrimental effect imposed by class imbalance\footnote{Detailed results at the class-level performance are presented in Appendix~\ref{app-results}.}. 
GCN$_\text{us, rw, st}$ improve the performance over GCN on all datasets, which demonstrate their generalizability and effectiveness in handling the imbalance issue. Specifically, GCN$_\text{us}$ and GCN$_\text{rw}$ share similar performance since essentially both of them balance the training loss while the performance of GCN$_\text{st}$ is unstable since randomly selecting linear interpolation coefficient cannot guarantee the generated instances follow the ground-truth class distributions. GraphSMOTE performs better than other baselines on datasets with higher homophily such as Cora, Cora-ML and DBLP, while worse on datasets with lower homophily such as Citeseer and Twitch PT. This is because edges in lower homophily networks tend to link nodes from different classes and therefore the pre-trained edge generator is also guided to link nodes from different classes. In this case, aggregating neighborhood features incorporates more noise information and therefore compromise the downstream classification. Despite that two advanced models, RECT and DR-GCN, achieve performance gain over the GCN baseline, both of their improvements are surprisingly no obvious compared with other baselines. For example, RECT achieves comparable performance to GCN on Cora and DBLP. For RECT, we hypothesize that the embeddings obtained by GNNs already encode the structural information and therefore further optimizing based on node proximity derives limited 
useful supervision. Another potential reason is because RECT is proposed for completely-imbalanced label while here we still have some 
training instances for each minority class. For DR-GCN, the reason of no significant performance improvement is that the limited training nodes may cause over-fitting of their generator and discriminator such that the class distribution is ill-defined via adversarial training~\cite{dong2019margingan}.

Moreover, we investigate the stability of the performance improvement achieved by our model by varying the imbalance ratio from 1:1 to 1:10. Specifically, we fix the number of training nodes as 2 for minoirty classes and gradually increasing the number of training nodes from 2 to 40 for majority classes, which exhausts the imbalance scenarios from being balanced to the heavily imbalanced. In Figure~\ref{fig-imbratio}, we can see DPGCN always performs the best when imbalance ratio is beyond 1:4 on the shown datasets and the gap grows larger as imbalance ratio further increases, which demonstrate the superority of DPGCN over other baselines in handling heavily imbalanced data. Besides, we observe that our model also achieves higher performance when data is balanced (imbalance ratio 1:1) on most datasets. This is because the imbalanced label propagation derives extra training supervision from pseudo labeled data and indicates the potential of our model in the balanced data setting. On Citeseer, our model achieves lower performance on the first few imbalance ratios because the inefficiency of imbalanced label propagation and smoothing of adjacent nodes due to lower homophily (i.e., 0.74) of Citeseer compared with other four datasets.

\vspace{-1.75ex}
\subsection{Ablation Study}
\begin{figure}[t]
     \centering
     \hspace{-1ex}
     \includegraphics[width=.46\textwidth]{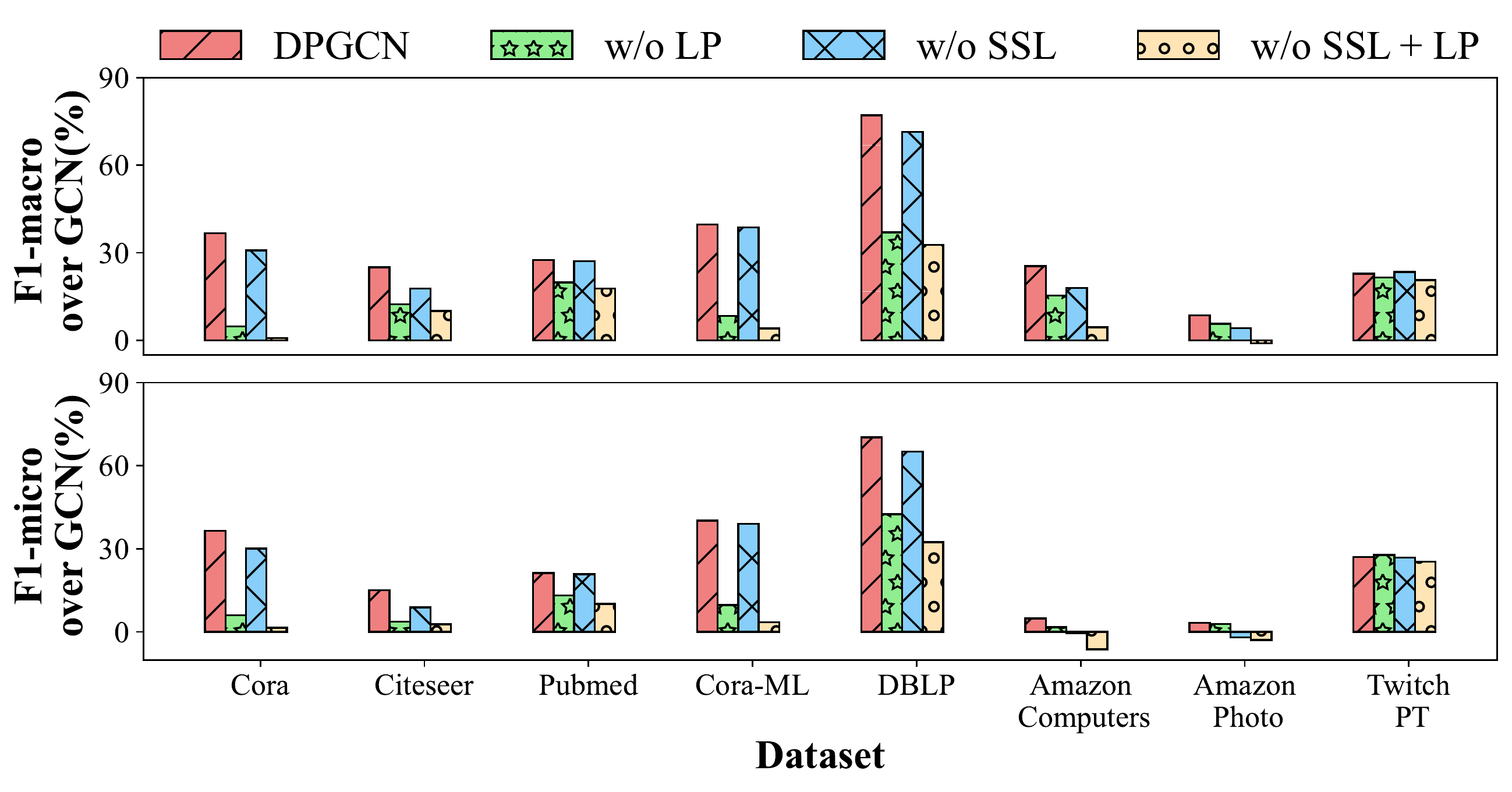}
     \vskip -3ex
     \caption{Ablation study results of DPGCN on all datasets.}
     \label{fig-ablation}
    \vskip -3ex
\end{figure}

To demonstrate the contribution of each proposed component in our framework, we conduct ablation studies and answer the second question in this subsection. In Figure~\ref{fig-ablation}, we present the performance improvement over GCN achieved by our proposed framework (\textbf{DPGCN}) along with variants that remove the imbalanced label propagation (\textbf{w/o LP)}, remove the self-supervised learning components (\textbf{w/o SSL}), and remove both (\textbf{w/o LP, SSL}). We use the best performing hyperparameters found for the results in Table~\ref{table-F1macro} and report the average F1-macro and F1-micro of 20 runs. Note that the most basic model here without SSL and LP has only episodic training with distance metric learning. Firstly, we observe that every proposed component contributes to the increase of F1-macro, which suggests the validity of each component in tackling class imbalance. However, performance becomes worse than GCN when solely applying episodic training with distance metric learning (yellow dot bar) on two Amazon Computers and Photos. This is because these two datasets have more classes and therefore the dimension of the concatenated embedding difference $\mathbf{g}$ in Eq.~\eqref{eq-concat} becomes pretty high and causes the effect of the curse of high dimensionality~\cite{aggarwal2001surprising}.

\vspace{-1ex}
\subsection{Influence of Threshold $\eta$}\label{sec-etatune}
\begin{figure}[t]
\vskip -1ex
     \centering
     \begin{subfigure}[b]{0.23\textwidth}
         \centering
         \includegraphics[width=0.97\textwidth]{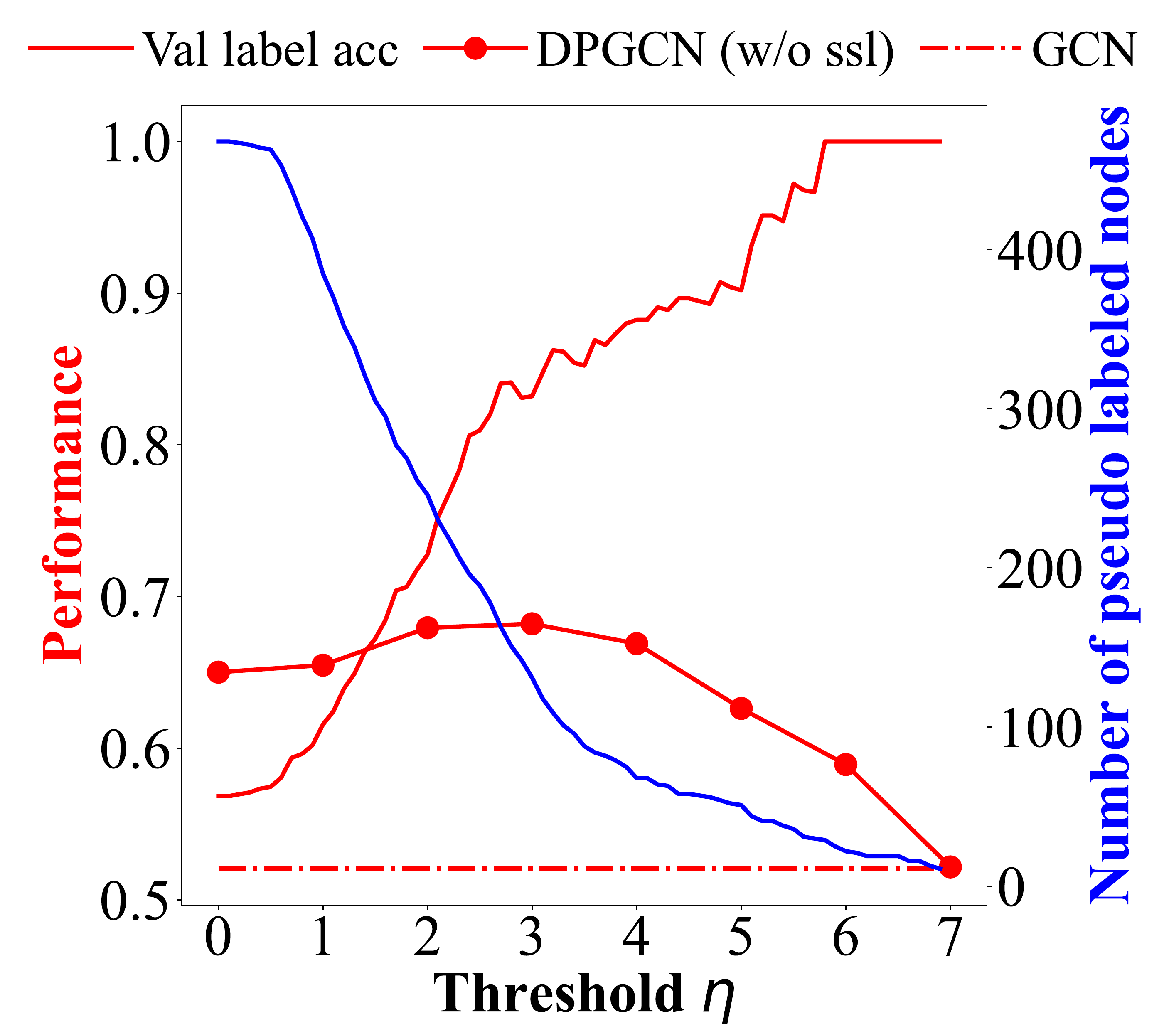}
         \vskip -1ex
         \caption{Cora}
         \label{fig-labeleffectcora}
     \end{subfigure}
     \begin{subfigure}[b]{0.23\textwidth}
         \centering
         \includegraphics[width=0.97\textwidth]{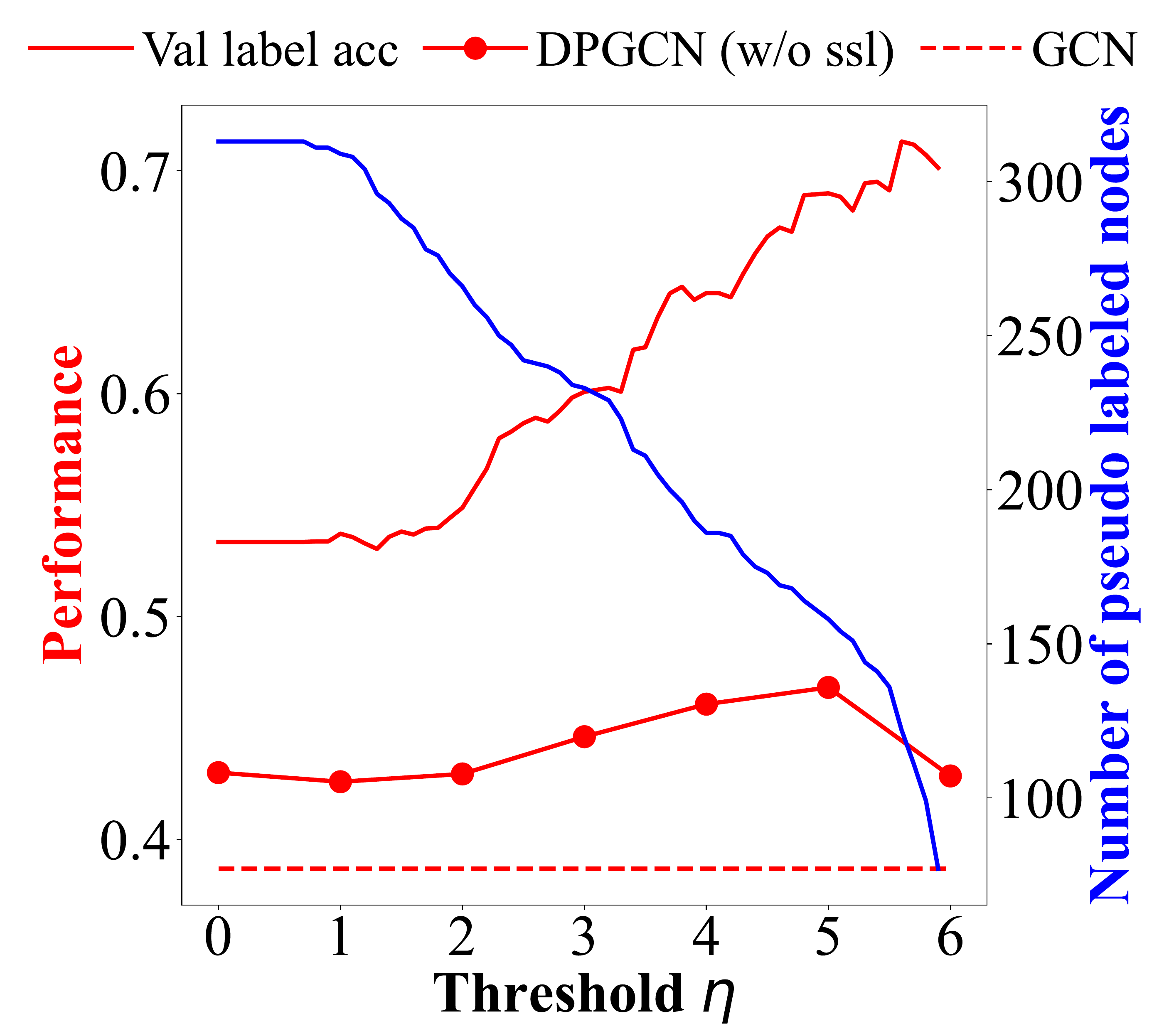}
         \vskip -1ex
         \caption{Citeseer}
         \label{fig-labeleffectciteseer}
     \end{subfigure}
     \vskip -2.5ex
     \caption{F1-macro score of using different threshold $\eta$ in imbalanced label propagation on Cora and Citeseer.}
     \label{fig-label3}
     \vskip -3ex
\end{figure}

Next we answer the last question by analyzing the influence of threshold $\eta$ in Eq.~\eqref{eq-threshold} on the model performance. Here we rule out the SSL component and only consider the performance of DPGCN (w/o SSL), which can exclusively reflect the effect of threshold $\eta$ in imbalanced label propagation. By varying $\eta$ from 0 to 7/6 on Cora/Citeseer, we obtain the total number of pseudo labels and their labeling accuracy from Eq.~\eqref{eq-threshold}. Meanwhile, we visualize the F1-macro score of DPGCN (w/o SSL) and the basic GCN model in Figure~\ref{fig-labeleffectcora}. In Cora, as $\eta$ increases the total number of pseudo labeled nodes decreases while the labeled quality (accuracy) increases since higher $\eta$ leads to precise labels with sharp distribution. In the initial stage when $\eta$ is between 0 and 1, the model performance stays the same because even though we obtain extra supervision from additional 500 pseudo labeled nodes as blue curve shows, only around 57\% of them are correctly classified and training with these low-quality data would compromise the model performance. As $\eta$ further rises from 1 to 3, the quality of pseudo labeled nodes improves since their accuracy increases from 0.57 to 0.86 during this stage, thus the F1-macro score of our DPGCN (w/o SSL) rises from 0.66 up to 0.68. After $\eta$ exceeds 3, even though the quality of the additional pseudo labels keep increasing, the model performance decreases because we have less pseudo labeled data. Finally, when $\eta$ is beyond 7 where no additional nodes other than the original imbalanced training nodes are labeled, DPGCN (w/o SSL) degenerate back to the DPGCN (w/o SSL, LP) and the performance gain achieved at this point is mainly due to 
class prototype-driven balanced training 
with distance metric learning. In Figure~\ref{fig-labeleffectciteseer}, we observe the similar trend in Citeseer except that in the degenerating stage when $\eta$ is 6, we still have performance gain over GCN while in Cora, the performance of DPGCN (w/o SSL, LP) converges to GCN, which is consistent with the previous ablation study in Figure~\ref{fig-ablation}. Note that here the label accuracy is solely computed for validation nodes by comparing their pseudo labels and ground-truth labels, which can also be treated as a hyperparameter tuning strategy for $\eta$. 

\vspace{-1ex}
\subsection{Further Explorations}\label{sec-further}
\begin{figure}[t]
     \centering
     \hspace{-3ex}
     \begin{subfigure}[b]{0.245\textwidth}
         \centering
         \includegraphics[width=0.97\textwidth]{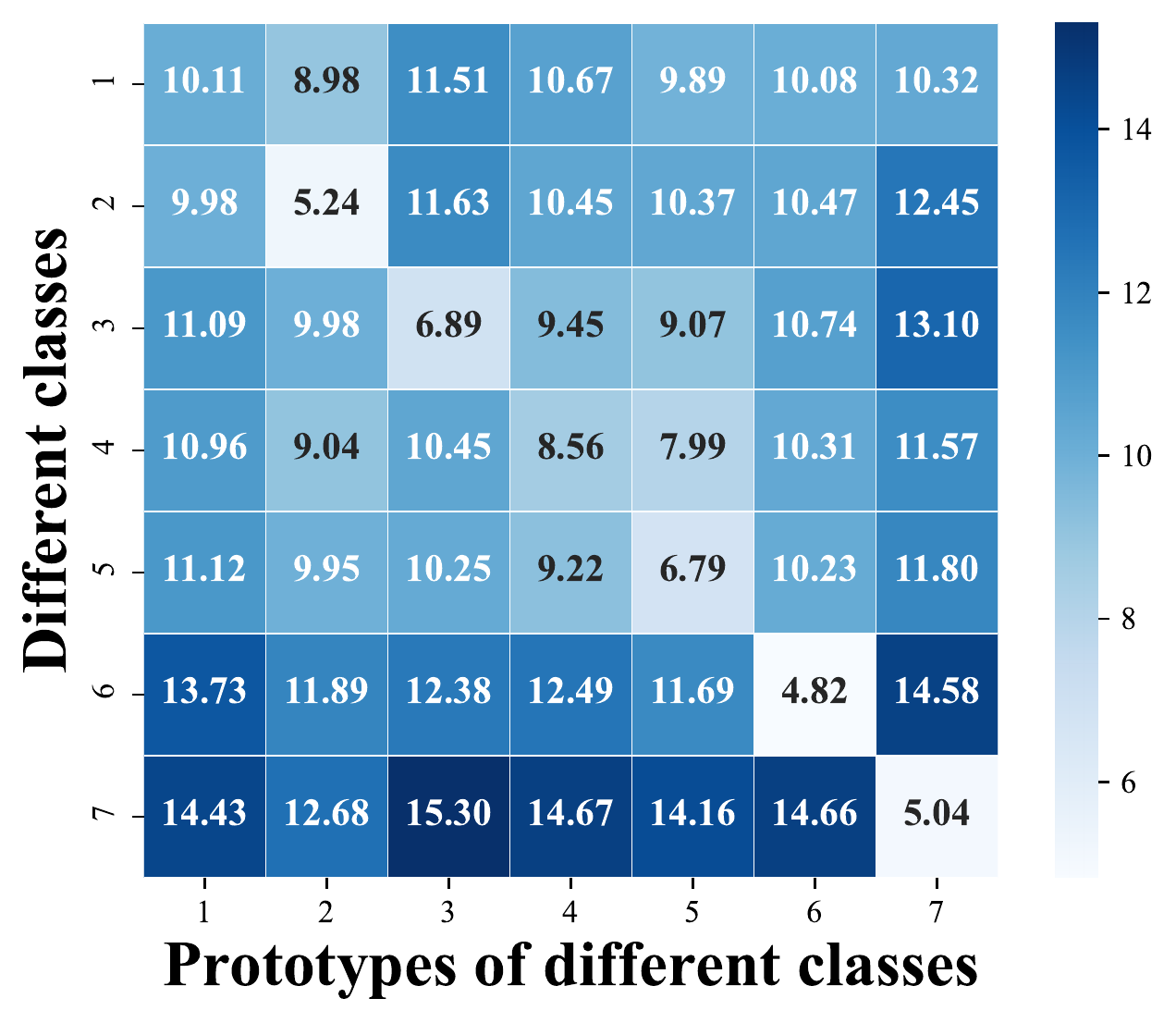}
         \label{fig-heatgcn}
     \end{subfigure}
     \hspace{-2ex}
     \begin{subfigure}[b]{0.245\textwidth}
         \centering
         \includegraphics[width=0.97\textwidth]{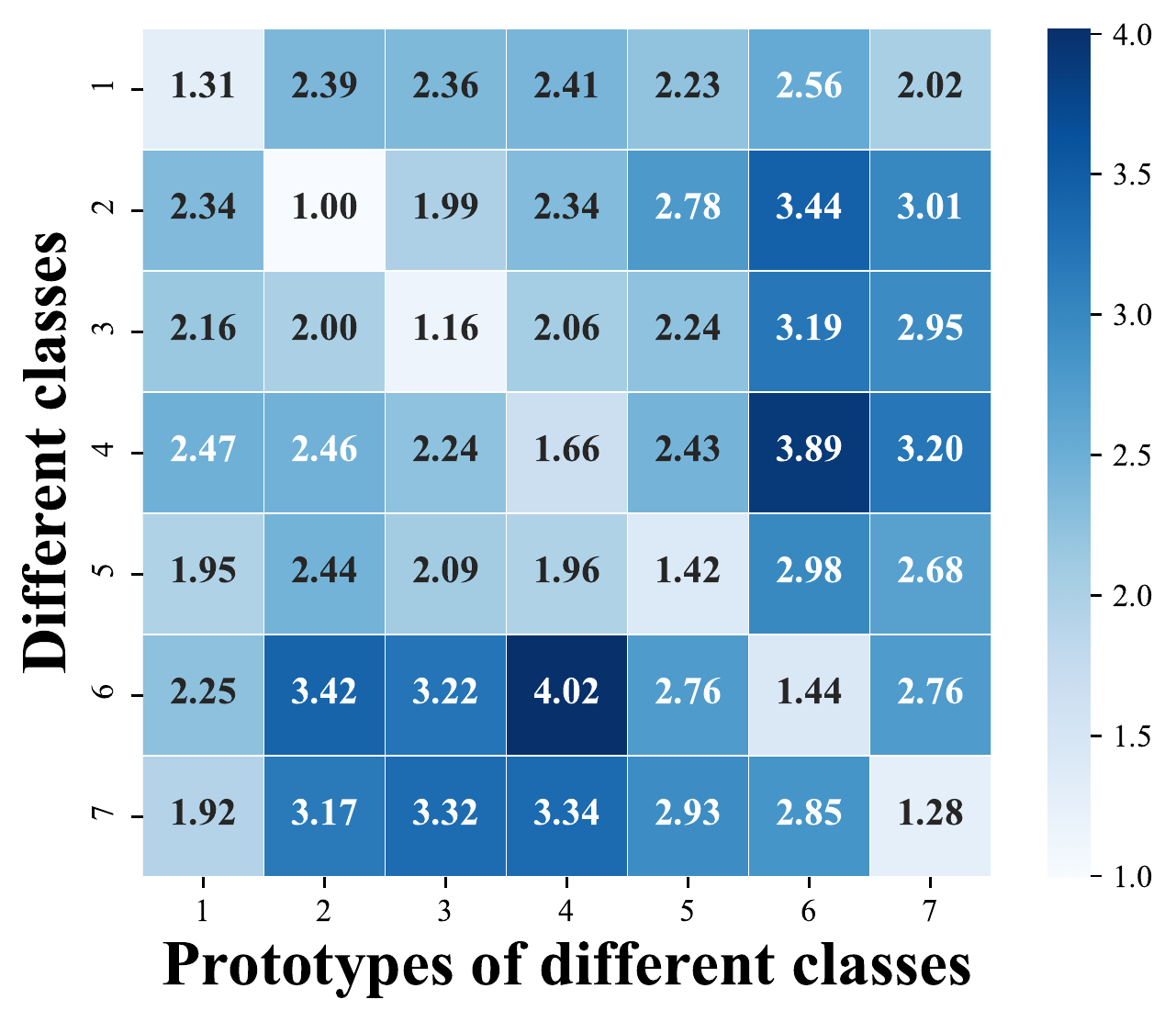}
         \label{fig-heatdpgcn}
     \end{subfigure}
     
     \hspace{-3ex}
     \begin{subfigure}[b]{0.245\textwidth}
         \centering
         \includegraphics[width=0.97\textwidth]{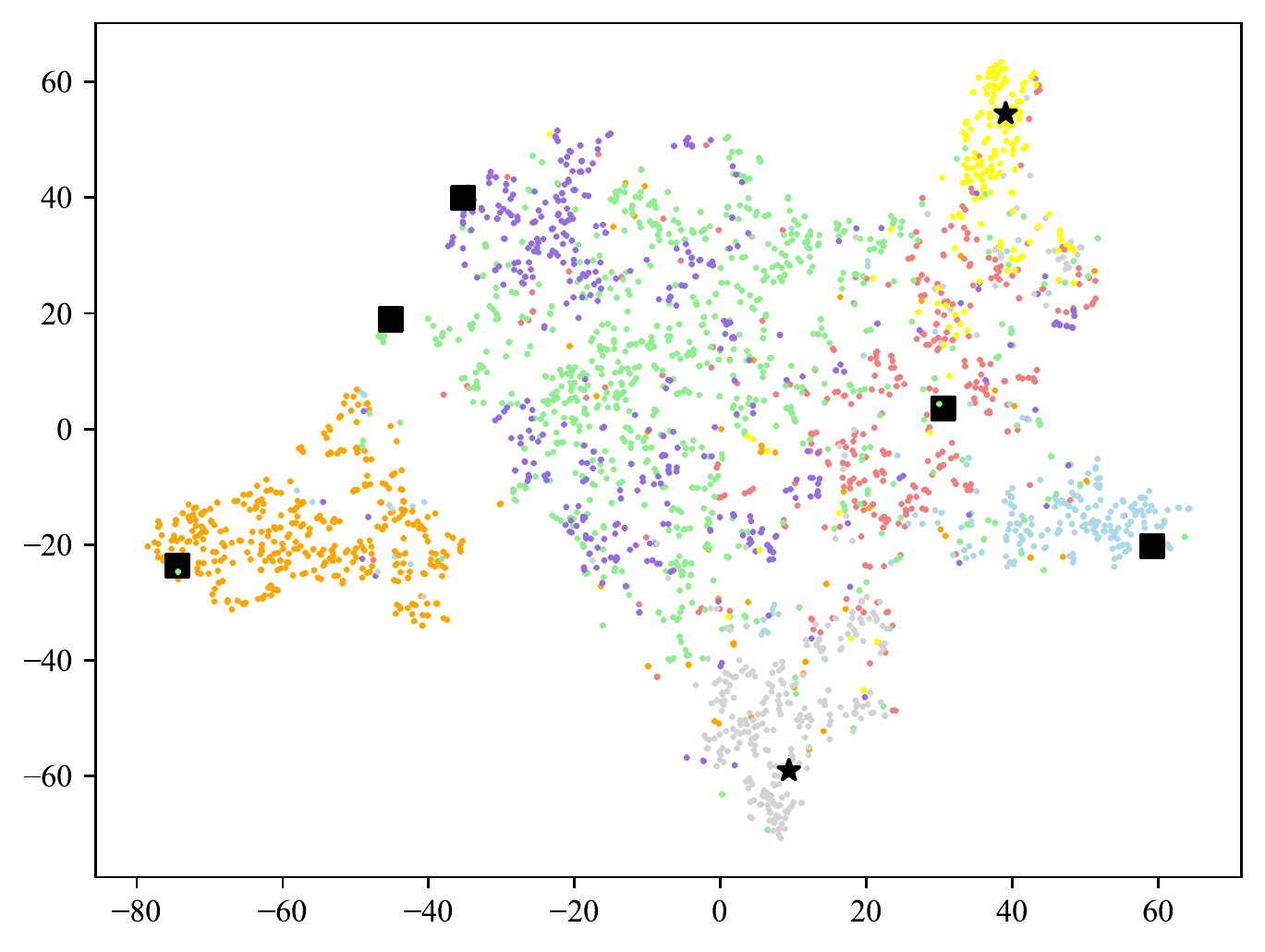}
         \caption{GCN}
         \label{fig-tsnegcn}
     \end{subfigure}
     \hspace{-1ex}
     \begin{subfigure}[b]{0.245\textwidth}
         \centering
         \includegraphics[width=0.97\textwidth]{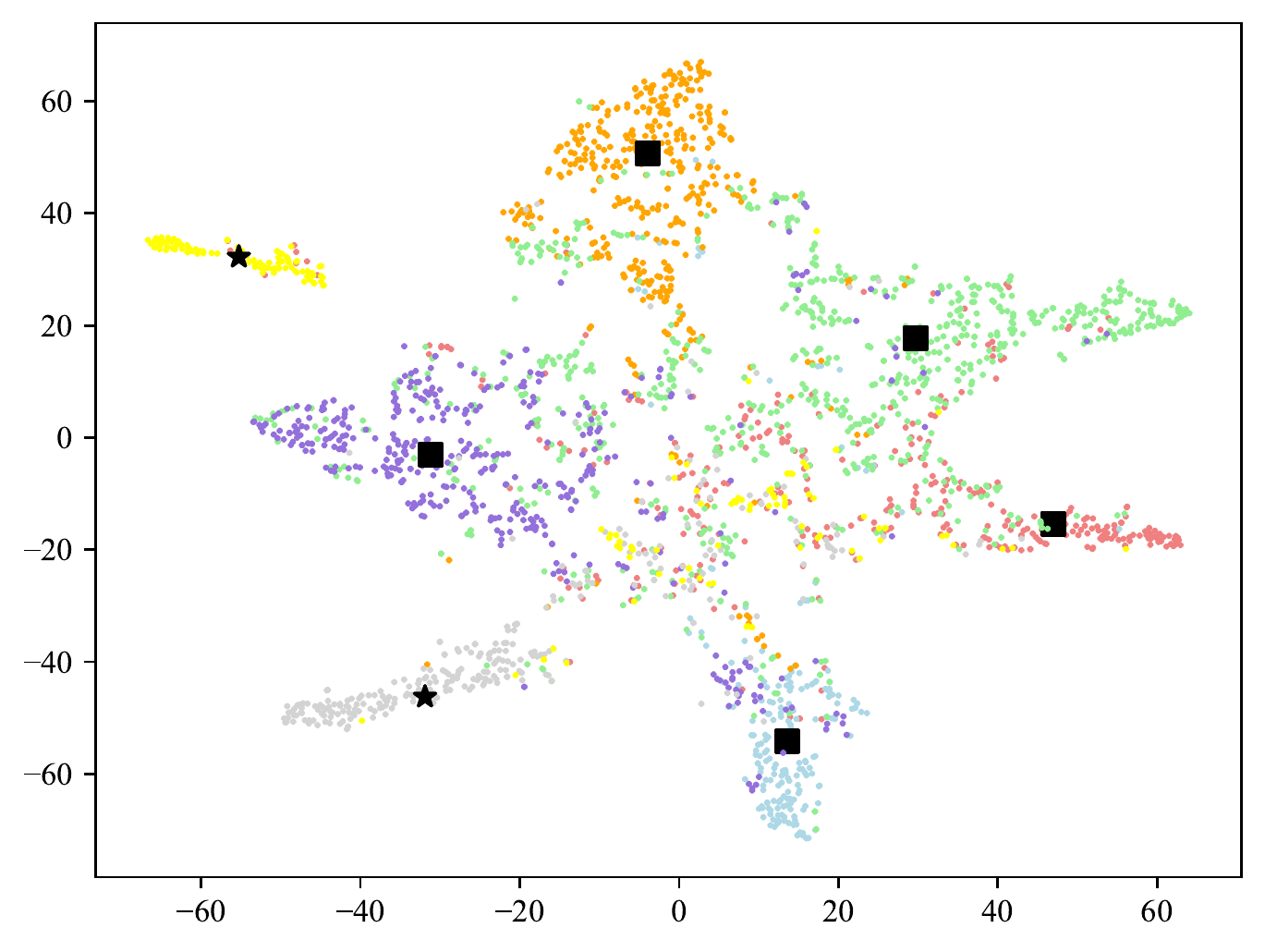}
         \caption{DPGCN}
         \label{fig-tsnedpgcn}
     \end{subfigure}
     \vskip -1.5ex
     \caption{
     Visualizing two comparisons of node embeddings by GCN to node distance metric embeddings by DPGCN. On top, the heatmap of average distance from nodes of each class to each class prototype, and on bottom, t-SNE, with the first five classes as minority and shown as black squares.
     }
     \vspace{-3ex}
     \label{fig-probe}
\end{figure}

Here we verify the effect of distance metric representation learning by calculating the average distance of node embeddings (distance metric representation) of each class to each class prototype and visualizing their corresponding embeddings by t-SNE~\cite{van2008visualizing} visualization in Figure~\ref{fig-probe}. The key difference in computing the value between the two heatmaps is that the first heatmap utilizes the averaged distance between embeddings of node and class prototypes (from GCN) while the second utilizes the averaged distance between distance metric representations between nodes and class prototypes. We can clearly see that directly utilizing the distance between embeddings fail to guarantee the shortest distance on the diagonal compared to all other ones in each row. For example, the distance from nodes in class 1 to the prototype of class 1, 10.11, is larger than the distance to the prototype class 2, 8.98, which could potentially lead to misclassification of nodes in class 1 into class 2. In contrast, the distance of distance metric representation on the diagonal is always the shortest one among each row, which therefore is better in classifying nodes based on their nearest class prototype. Note that most of the distance values in the first heatmap are larger than the second one since we smooth the distance metric representation of connected nodes through self-supervised learning, which pulls distance metric representation of nodes closer as a whole. The second row of Figure~\ref{fig-probe} further demonstrates the t-SNE visualization of embeddings and distance metric representations of nodes in different classes along with their class prototypes. The node embeddings obtained from GCN has the Silhouette score~\cite{rousseeuw1987silhouettes} 0.099 while the node distance metric representation from DPGCN has the score 0.117, which indicates the node distance metric representation is more well-apart from each other and thus is more distinguished. Besides, for majority class, the embeddings and the distance metric representation cluster tightly around corresponding prototypes, such as grey and yellow, while they cluster loosely or even falsely cluster around other different class prototype and thus are hard to be classified for minority classes, such as green. This is because minority class has less training nodes and thus the computed class prototypes by mean-pooling are biased only towards few given training nodes and fail to be representative of the whole class. We observe that such bad clustering is mitigated after applying DPGCN and utilizing distance metric representation.

\section{Conclusion}\label{sec-conclusion}
In this paper, we focused on the imbalanced node classification problem in graphs, which widely exists in real-world setting such as malicious user detection and drug function testing. Noticing that imbalanced node classification naturally inherits issues of deep learning for imbalance classification: the inclination to learning towards majority classes and the catastrophic forgetting of previous learned instances in minority classes, we use a class prototype-driven balance training scheme to balance the training loss between different classes. To further alleviate the issue of unrepresentative class prototypes, we construct another distance metric space to fully leverage the distance information of query nodes to all class prototypes. Moreover, we derive extra supervision from network topology by an imbalanced label propagation scheme and smooth the learned distance metric representation between adjacent nodes. Experiments on 8 real-world datasets demonstrate the effectiveness of the proposed DPGNN framework in relieving the class imbalance issue. For future work, we plan to study imbalanced graph classification and expect to utilize the graph topology as extra supervision.


\balance
\bibliographystyle{ACM-Reference-Format}
\bibliography{references}









\clearpage
\newpage
\appendix
\section{Notations}\label{notation}
\begin{table}[h!]
\setlength{\tabcolsep}{2pt}
\setlength{\extrarowheight}{.09pt}
\caption{Table of main symbols.}
\centering
\vspace{-2ex}
\begin{tabular}{c@{\hskip 2ex}l}
\Xhline{2\arrayrulewidth} 
\textbf{Symbols} & \textbf{Definitions}\\
 \Xhline{2\arrayrulewidth} 
 $\mathcal{G}$ & unweighted and undirected attributed network\\
 $\mathcal{V}$ & set of nodes in the network\\
 $\mathcal{V}_l$ & set of labeled nodes in the network\\
 $\mathcal{E}$ & set of edges in the network\\
 $\mathbf{A}$ & adjacency matrix\\
 $\mathbf{X}$ & node attribute matrix\\
 $\mathbf{Y}$ & one-hot node label matrix\\
$\mathcal{C}$ & set of node classes in the network\\
$\mathcal{N}_{v_i}$ & neighborhood node set of node $v_i$ \\
 $\mathcal{S}_c$ & support set of class $c$ \\ 
 $\mathcal{Q}_c$ & query set of class $c$ \\ 
 $\mathbf{H}$ & node representation matrix\\
 $\mathbf{H}^{\mathcal{S}_c}$ & representation matrix of nodes in support set $\mathcal{S}_c$ \\
 $\mathbf{H}^{\mathcal{Q}_c}$ & representation matrix of nodes in query set $\mathcal{Q}_c$  \\
 $\mathbf{P}$ & prototypical representation matrix\\
 $\mathbf{G}$ & distance representation matrix\\
  $\widetilde{\mathbf{Y}}$ & reweighted node label matrix\\
 $\widehat{\mathbf{Y}}$ & propagated label disbribution\\
 $\mathbf{t}$ & topological information gain\\
 $\check{\mathbf{Y}}$ & one-hot pseudo labeled matrix\\
 $f_1$ & GNN-based encoder to extract node representation\\
 $f_2$ & linear layer to extract distance metric representation\\
 $\chi$ & edge homophily\\
 $\eta$ & threshold in imbalanced label propagation\\
 
 \Xhline{2\arrayrulewidth}
\end{tabular}
\label{tab-notation}
\vskip -4ex
\end{table}

\section{Edge Homophily}\label{sec-homophily}
A key assumption in most graph-based prediction algorithm such as Label Propagation and Graph Neural Network is the network homophily. It measures the feature/class similarity level of adjacent nodes in the graph. Following the literature~\cite{zhu2020beyond}, we use the edge homophily as the metric here and the computation is:
\begin{equation}
    \chi^{\mathcal{G}} = \frac{|\{(v_i, v_j)\in\mathcal{E}^{\mathcal{G}}: \mathbf{Y}_i = \mathbf{Y}_j\}|}{|\mathcal{E}^{\mathcal{G}}|},
\end{equation}
where high $\chi^{\mathcal{G}}$ means many edges connect nodes of the same class or sharing the similar feature distributions.

\section{Hyperparameter tunning}\label{app-hyper}
For the encoder $f_1$, we stack two GCN layers with the ReLU activation function and set the number of hidden units to be 256. All hyperparameters are tuned on the validation set. The tuning range of hyperparameters is as follows:
\begin{itemize}
\item \textbf{Early stopping:} $0$
    \item \textbf{Dropout:}$\{0, 0.5\}$
    \item \textbf{Weight decay:} $1^{\text{st}}$ layer, $5e^{-4}$; $2^{\text{nd}}$ layer, $0$
    \item \textbf{Training coefficients:} $\lambda_1, \lambda_2 \in \{1, 10, 20\}$
    \item \textbf{Threshold:} $\eta \in \{1, 2, 3, 4, 5, 6\}$
    \item \textbf{Training epoch:} $\{1000, 3000, 6000\}$
    \item \textbf{Label propagation epoch:} $\{10, 20\}$
    
\end{itemize}

\section{Results}\label{app-results}
\subsection{Performance Comparison}\label{app-compare}
From the following tables, we could easily see that the performance of minority classes lower than majority classes. However, an interesting observation is that the performance is also different across different classes, which indicates that the same imbalance ratio may impose different level of negative influence on different classes. We hypothesize that this is due to the graph topology. Since prediction in graph neural network is made after propagation, it is highly possible that nodes in different minority classes own different influence on other nodes in this graph. Therefore, even though different minority classes share the same quantity-based imbalance ratio, they may still have different topology and we term it as topology imbalance. Such topology imbalance only exists in graph-structured data and is almost unexplored yet. We include it as one future work.
\begin{table}[h!]
\vskip -2.5ex
\tiny
\setlength{\extrarowheight}{.095pt}
\setlength\tabcolsep{3pt}
\centering
\caption{Node classification performance on Cora.}
\vspace{-4ex}
\begin{tabular}{|l|ccccc|cc|ccc|}
\Xhline{2\arrayrulewidth}
\multirow{2}{*}{Model} & \multicolumn{7}{c|}{Classes (2:2:2:2:2:20:20)} & \multicolumn{3}{c|}{Performance} \\
\cline{2-11}
 & \textbf{1} & \textbf{2} & \textbf{3} & \textbf{4} & \textbf{5} & 6 & 7 & F1-macro & F1-weight & F1-micro \\
 \hline
GCN & 0.3115 & 0.6274 & 0.7327 & 0.4903 & 0.4973 & 0.5259 & 0.4582 & 0.5205 & 0.5195 & 0.5212 \\
GCN-us & 0.3361 & 0.6157 & 0.7806 & 0.5723 & 0.4930 & 0.5698 & 0.5744 & 0.5631 & 0.5659 & 0.5727 \\
GCN-rw & 0.3334 & 0.6297 & 0.7757 & 0.5830 & 0.4904 & 0.5544 & 0.5598 & 0.5609 & 0.5660 & 0.5724 \\
GCN-st & 0.3278 & 0.6011 & 0.7702 & 0.5086 & 0.4693 & 0.5929 & 0.5716 & 0.5488 & 0.5398 & 0.5519 \\
GraphSMOTE & 0.4333 & 0.6780 & 0.8215 & 0.5880 & 0.6644 & 0.4971 & 0.4091 & 0.5845 & 0.6026 & 0.5820 \\
RECT &  0.4653 & 0.5920 & 0.6542 & 0.4404 & 0.3957 & 0.5331 & 0.5828 & 0.5234 & 0.5025 & 0.5448  \\
DR-GCN & 0.1722 & 0.7486 & 0.8446 & 0.4927 & 0.5244 & 0.4585 & 0.6180 & 0.5513 & 0.5362 & 0.5520 \\
DPGCN & 0.5478 & 0.7238 & 0.8615 & 0.6410 & 0.7216 & 0.7675 & 0.7174 & 0.7115 & 0.7029 & 0.7111\\
\Xhline{2\arrayrulewidth}
\end{tabular}
\end{table}

\begin{table}[h!]
\vskip -2.5ex
\tiny
\setlength{\extrarowheight}{.095pt}
\setlength\tabcolsep{3pt}
\centering
\caption{Node classification performance on Citeseer.}
\vspace{-4ex}
\begin{tabular}{|l|cccc|cc|ccc|}
\Xhline{2\arrayrulewidth}
\multirow{2}{*}{Model} & \multicolumn{6}{c|}{Classes (2:2:2:2:20:20)} & \multicolumn{3}{c|}{Performance} \\
\cline{2-10}
 & \textbf{1} & \textbf{2} & \textbf{3} & \textbf{4} & 5 & 6 & F1-macro & F1-weight & F1-micro \\
 \hline
GCN & 0.0587 & 0.3168 & 0.3772 & 0.4086 & 0.6424 & 0.5182 & 0.3870 & 0.4169 & 0.4692\\
GCN-us & 0.0977 & 0.3716 & 0.4306 & 0.4865 & 0.7147 & 0.6007 & 0.4503 & 0.4822 & 0.5220 \\
GCN-rw & 0.0864 & 0.3327 & 0.4617 & 0.5383 & 0.6572 & 0.5979 & 0.4457 & 0.4800 & 0.5156 \\
GCN-st & 0.0891 & 0.3451 & 0.4599 & 0.5064 & 0.6907 & 0.5861 & 0.4462 & 0.4794 & 0.5127 \\
GraphSMOTE & 0.0247 & 0.2329 & 0.6394 & 0.6207 & 0.5027 & 0.5212 & 0.4236 & 0.4774 &0.5020   \\
RECT & 0.1181 & 0.2454 & 0.3618 & 0.4355 & 0.6687 & 0.5719 & 0.4002 & 0.4243 & 0.4549  \\
DR-GCN & 0.0526 & 0.0471 & 0.6719 & 0.3937 & 0.6358 & 0.5532 & 0.3924 & 0.4414 & 0.4880  \\
DPGCN & 0.1295 & 0.3675 & 0.5459 & 0.5232 & 0.6895 & 0.6474 & 0.4838 & 0.5180 & 0.5397\\
\Xhline{2\arrayrulewidth}
\end{tabular}
\end{table}


\begin{table}[h!]
\vskip -2.5ex
\tiny
\setlength{\extrarowheight}{.095pt}
\setlength\tabcolsep{3pt}
\centering
\caption{Node classification performance on Cora-ML.}
\vspace{-4ex}
\begin{tabular}{|l|ccccc|cc|ccc|}
\Xhline{2\arrayrulewidth}
\multirow{2}{*}{Model} & \multicolumn{7}{c|}{Classes (2:2:2:2:2:20:20)} & \multicolumn{3}{c|}{Performance} \\
\cline{2-11}
 & \textbf{1} & \textbf{2} & \textbf{3} & \textbf{4} & \textbf{5} & 6 & 7 & F1-macro & F1-weight & F1-micro \\
 \hline
GCN & 0.3114 & 0.6274 & 0.7327 & 0.4903 & 0.4973 & 0.5259 & 0.4582 & 0.5205 & 0.5195 & 0.5212 \\
GCN-us & 0.5659 & 0.4490 & 0.7155 & 0.4704 & 0.4931 & 0.5549 & 0.7106 & 0.5656 & 0.5516 & 0.5611 \\
GCN-rw & 0.3334 & 0.6297 & 0.7757 & 0.5830 & 0.4904 & 0.5544 & 0.5598 & 0.5609 & 0.5660 & 0.5724 \\
GCN-st & 0.3278 & 0.6011 & 0.7702 & 0.5086 & 0.4693 & 0.5929 & 0.5716 & 0.5488 & 0.5398 & 0.5519 \\
GraphSMOTE & 0.5957 & 0.5742 & 0.7739 & 0.7051 & 0.6295 & 0.2979 & 0.7870 & 0.6233 & 0.6450 & 0.6130 \\
RECT & 0.6825 & 0.3485 & 0.7085 & 0.3479 & 0.5681 & 0.4731 & 0.7427 & 0.5530 & 0.5560 & 0.6026 \\
DR-GCN & 0.6734 & 0.4813 & 0.8961 & 0.3532 & 0.1598 & 0.5604 & 0.6643 & 0.5412 & 0.4716 & 0.5060 \\
DPGCN & 0.7539 & 0.5746 & 0.9026 & 0.6705 & 0.7064 & 0.6752 & 0.8080 & 0.7273 & 0.7278 & 0.7305\\
\Xhline{2\arrayrulewidth}
\end{tabular}
\end{table}

\begin{table}[h!]
\vskip -2ex
\tiny
\setlength{\extrarowheight}{.095pt}
\setlength\tabcolsep{3pt}
\centering
\caption{Node classification performance on DBLP.}
\vspace{-4ex}
\begin{tabular}{|l|ccc|c|ccc|}
\Xhline{2\arrayrulewidth}
\multirow{2}{*}{Model} & \multicolumn{4}{c|}{Classes (2:2:2:20)} & \multicolumn{3}{c|}{Performance} \\
\cline{2-8}
 & \textbf{1} & \textbf{2} & \textbf{3} & 4 & F1-macro & F1-weight & F1-micro \\
 \hline
GCN & 0.5250 & 0.2319 & 0.3362 & 0.2997 & 0.3482 & 0.3829 & 0.3876\\
GCN-us & 0.6081 & 0.3083 & 0.4006 & 0.3688 & 0.4214 & 0.4599 & 0.4795 \\
GCN-rw & 0.6113 & 0.3392 & 0.4344 & 0.3667 & 0.4379 & 0.4744 & 0.4892\\
GCN-st & 0.6221 & 0.2413 & 0.3079 & 0.3316 & 0.3757 & 0.4302 & 0.4522\\
GraphSMOTE & 0.6339 & 0.3158 & 0.6949 & 0.2931 & 0.4844 & 0.4938 & 0.4530\\
RECT & 0.4042 & 0.3252 & 0.3015 & 0.3443 & 0.3438 & 0.3602 & 0.3810\\
DR-GCN & 0.6886 & 0.0583 & 0.4172 & 0.3546 & 0.3797 & 0.4190 & 0.4510\\
DPGCN & 0.7976 & 0.5715 & 0.6821 & 0.4158 & 0.6167 & 0.6665 & 0.6597\\
\Xhline{2\arrayrulewidth}
\end{tabular}
\end{table}

\begin{table}[h!]
\vskip -2.5ex
\tiny
\setlength{\extrarowheight}{.095pt}
\setlength\tabcolsep{3pt}
\centering
\caption{Node classification performance on Twitch-PT.}
\vspace{-4ex}
\begin{tabular}{|l|c|c|ccc|}
\Xhline{2\arrayrulewidth}
\multirow{2}{*}{Model} & \multicolumn{2}{c|}{Classes (2:20)} & \multicolumn{3}{c|}{Performance} \\
\cline{2-6}
 & \textbf{1} & 2 & F1-macro & F1-weight & F1-micro \\
 \hline
GCN & 0.4408 & 0.4706 & 0.4557 & 0.4510 & 0.4656 \\
GCN-us & 0.5454 & 0.4381 & 0.4917 & 0.5088 & 0.5131 \\
GCN-rw & 0.5602 & 0.4323 & 0.4963 & 0.5168 & 0.5193\\
GCN-st & 0.5817 & 0.4187 & 0.5002 & 0.5267 & 0.5301\\
GraphSMOTE & 0.2798 & 0.5046 & 0.3922 & 0.3558 & 0.4130\\
RECT & 0.4729 & 0.5068 & 0.4898 & 0.4843 & 0.4904\\
DR-GCN & 0.7026 & 0.2557 & 0.4791 & 0.5524 & 0.5750\\
DPGCN & 0.6661 & 0.4539 & 0.5600 & 0.5944 & 0.5915\\
\Xhline{2\arrayrulewidth}
\end{tabular}
\end{table}

\end{document}